\documentclass[conference]{IEEEtran}
\ifCLASSINFOpdf
\else
\fi
\usepackage{amssymb}
\usepackage{amsmath,bm}
\usepackage{algorithm}
\usepackage{algorithmic}
\newcommand{\PCR}{\mathrm{PCR6}}
\newcommand{\BetP}{\mathrm{BetP}}

\newcommand{\LNS}{\mathrm{LNS}}

\newcommand{\conj}{\mathrm{conj}}

\newcommand{\DP}{\mathrm{DP}}
\newcommand{\DS}{\mathrm{Dempster}}

\newcommand{\caut}{\mathrm{Cautious}}
\newcommand{\conf}{\mathrm{Conf}}
\newcommand{\citet}{\cite}
\newcommand{\citep}{\cite}

\newcommand{\ocap}{\operatornamewithlimits{\text{\textcircled{\scalebox{1.4}{\tiny{$\cap$}}}}}}

\newcounter{ExpNum}
\usecounter{ExpNum}
\newcommand{\Exp}{\noindent\textbf{Experiment~\arabic{ExpNum}}\refstepcounter{ExpNum}}
\usepackage{graphicx}
\usepackage{slashbox,pict2e}
\begin{document}

\title{Evidence combination for a large number of sources}
\author{\IEEEauthorblockN{Kuang Zhou$^\text{a}$,
Arnaud Martin$^\text{b}$, and
Quan Pan$^\text{a}$
}
\IEEEauthorblockA{a.  Northwestern Polytechnical University,
Xi'an, Shaanxi 710072, PR China. }
\IEEEauthorblockA{b. DRUID, IRISA, University of Rennes 1, Rue E. Branly, 22300 Lannion, France}
}

\maketitle

\begin{abstract}
The theory of belief functions is an effective tool to deal with the  multiple uncertain
information. In recent years, many evidence combination rules have been proposed in this framework, such as
the conjunctive rule, the cautious rule, the PCR (Proportional Conflict Redistribution)
rules and so on. These rules can be adopted for different types
of sources. However, most of these rules are not applicable when the number of sources is large. This is
due to either the complexity  or the  existence of an absorbing
element (such as the total conflict mass function for the conjunctive-based rules when applied on unreliable evidence).
In this paper, based on the assumption that the majority of sources are reliable,
a combination rule for a large number of sources, named LNS (stands for Large Number of Sources), is proposed on the basis of
a simple idea: the more common ideas one source shares with others, the more
reliable the source is. This rule is adaptable for aggregating  a large number of sources among which some are unreliable.
It will keep the spirit of the conjunctive rule to reinforce the belief on the focal elements with
which the sources are in agreement. The mass on the empty set will  be kept as an indicator of the conflict. Moreover, it
can be used to elicit the major opinion among the experts. The experimental results on synthetic mass functions
verify that  the rule  can be effectively used to combine a large number of mass functions and to
elicit the major opinion.
\end{abstract}


\IEEEpeerreviewmaketitle

\section{Introduction}
The theory of belief functions (also called Dempster--Shafer Theory, DST) provides  effective tools to modele  the
uncertain information and  to combine them using a combination rule.
One of the classical combination rule in the belief function framework is the Dempster's rule~\cite{Smets07a}. But
this rule has been criticized because of its unexpected behavior under some situations
enlightened by the famous Zadeh's example. 
Smets proposed a modification of Dempster's rule,  often called ``conjunctive rule", where
the empty set can be assigned with
a non-null mass under the Transferable
Belief Model (TBM) \citep{smets1990combination}. 
In fact the conjunctive rule is equivalent to the Dempster rule  without the normalization
process. It has a fast and clear convergence towards a solution.

Smets supports that the mass on the empty set, called also global conflict, can play a role of alarm \citep{smets1990combination}.
When the global conflict is
high, it indicates that there is  strong disagreement among the sources of mass functions to be fused.  However, as
observed in \cite{martin2008conflict,destercke2013toward}, the mass on the empty set is not sufficient to exactly describe the
conflict since it includes an amount of auto-conflict \cite{Martin06a}. Even if the sources are reliable, they can be of high
conflict in sense of the mass assigned to the empty set.
When using the conjunctive rule, even if there is only
a small amount of concordant evidence, the total conflict mass function, {\em i.e.} $m(\emptyset)=1$ will be an absorbing element.
Consequently, when combining a large number of (incompatible) mass functions using the conjunctive
rule, the global conflict  may tend to 1. This makes it impossible to reveal the
cause of high global conflict. We do not know whether it is due
to the  sources to fuse or  caused by the absorption power of the empty set \cite{martin2008conflict}. In other words, even
the combined mass function by the conjunctive rule is $m(\emptyset)\approx 1$, the proposition that the sources are highly conflicting may
be incorrect.

In order to  rectify the drawbacks of the classical Dempster's rule and conjunctive rule,
many approaches have been made through the modification of the combination rule. However, most of them are not efficient when applied on a large number of sources either due to the ineffective way to handle
conflict or the high complexity of the computation.
Moreover, the major opinion among all the participants is not easy to found.
In some applications such as crowdsourcing, there are usually a large number of sources. An important problem for crowdsourcing systems is
to identify the experts who tend to answer the questions correctly among participants. Finding the reliable workers
in the system can improve the quality of knowledge one can extract from crowds. One of the most commonly used assumptions in crowdsourcing systems
is that the majority of participants can give correct answers. Thus if the major opinion among the participants
could be elicited, the reliable workers can be easily found.

We propose in this paper, a conjunctive-based
combination rule, named $\LNS$ (stands for Large Number of Sources), in order  to
aggregate  a large number of mass functions. This rule is based on the idea that combining mass functions from different sources is
similar to combining opinions from multiple stake-holders in
group decision-making. 
Hence, the more one's opinion is consistent with the other experts, the more reliable the source is.
To build our rule, we assume that all the mass functions available are separable mass functions. A separable mass function can
be expressed by a group of simple support mass functions, such as non-dogmatic mass functions. In many applications, the mass
assignments are directly in the form of Simple Support Functions (SSF) \cite{ds2}.
Hence, we can group the SSFs in such a way that sources in the same group have the same focal elements or have focal element without conflict. Mass functions in each small group
are first fused and then discounted according to the proportions of SSFs in each group. After that the number of mass functions to fuse is the number of groups which is independent of the number of sources.
Therefore, the problem brought by the absorbing element (the empty set) using the conjunctive rule can be avoided.

The rest of this paper is organized as follows. In Section 2, some basic knowledge of belief function theory is briefly introduced. The proposed
evidence combination approach is presented in detail in Section 3.  Numerical examples are employed to compare different
combination rules and show the effectiveness of $\LNS$ rule in Section 4.
Finally, Section 5 concludes the paper.

\section{Background}
\subsection{Basic knowledge on the theory of belief functions}
We consider $\Theta=\{\theta_{1},\theta_{2},\ldots,\theta_{n}\}$ such as the discernment frame. A mass function is defined on the power
set $2^{\Theta}=\{A:A\subseteq\Theta\}$. The mass function $m:2^{\Theta}\rightarrow[0,1]$ is said to be a Basic Belief
Assignment (BBA) on $\text{2}^{\Theta}$, if it satisfies:
\begin{equation}
\sum_{A\subseteq\Theta}m(A)=1.
\end{equation}
If $m(A)>0$, with $A\in2^{\Theta}$, $A$ is called a focal element, and the set of focal elements is denoted by $\mathcal{F}$.

The frame of discernment $\Theta$ can also be a focal element. If $\Theta$ is a focal element, the mass function is called non-dogmatic. The mass assigned to the frame of discernment, $m(\Theta)$, is interpreted as a degree of ignorance. In the case of total ignorance, $m(\Theta)=1$. This mass function is also called a vacuous mass function.
If there is only one focal element, {\em i.e.}
$m(A)=1, A \subset \Theta$, the mass function is  categorical.  Another special case of assignment is named consonant mass functions, where the focal
elements include each other as a subset, {\em i.e.} if $A, B \in \mathcal{F}, A \subset B ~\text{or}~ B \subset A$.


In order to combine information sources assumed reliable and cognitively independent, the conjunctive combination is usually used, given by:
\begin{equation}
\label{conjunctive}
m_\conj(X) =(\ocap_{j=1, \cdots, S}m_j)(X)= \sum\limits_{Y_1 \cap \cdots \cap Y_S = X} \prod_{j=1}^{S}m_j(Y_j),
\end{equation}
where 
$m_j(Y_j)$ is the mass allocated to $Y_j$ by expert $j$. Another kind of conjunctive combination is Dempster's rule \cite{dempster1967upper} given by:
\begin{equation}
   m_\DS(X)= \begin{cases}
    0 & \text{if}~ X = \emptyset,\\
     \frac{m_\conj(X)}{1-m_\conj(\emptyset)} & \text{otherwise}.
  \end{cases}
\end{equation}
The item
$\kappa \triangleq m_\conj (\emptyset)$ 
is generally called Dempster's degree of conflict  of the combination or the inconsistency of the combination.

The disjunctive rule \cite{smets1993belief} can be used if we only assume that at least one of the sources is reliable.


If information sources are dependent the cautious rule \cite{denoeux2006cautious} can be applied. Cautious combination of $S$ non-dogmatic mass functions $m_j, j=1,2,\cdots,S$  is defined by the BBA with the following weight function:
\begin{equation}
  w(A)= \mathop{\wedge}\limits_{j=1}^S w_j(A), ~~ A \in 2^\Theta \setminus \Theta.
\end{equation}
We thus have
\begin{equation}
 m_\caut(X) = \ocap_{A \subsetneq \Theta} A^{\mathop{\wedge}\limits_{j=1}^S w_j(A)},
\end{equation}
where $A^{w_j(A)}$ is the simple support function focused on $A$ with weight function $w_j(A)$ issued from the canonical decomposition of $m_j$.
Note also that $\wedge$ is the min operator. 
Moreover, in the case of dependant sources, the average combination rule can be choosen.


In order to manage the $\kappa$ value by redistributing it on partial ignorance, the Dubois and Prade rule ($\DP$ rule) \cite{dubois1988representation}, can be applied. It can be seen as a mixed conjunctive and disjunctive rule.


Moreover the \textbf{$\PCR$} proposed by \citet{Martin06a} is one of the most popular rule to combine hight conflicting sources.

\subsection{Discounting process based on source-reliability}
\label{reliability}

When the sources of evidence are not completely reliable, a discounting operation proposed by~\citet{ds2} 
can be applied. Denote the reliability degree of mass function $m$ by $\alpha \in[0,1]$, then the discounting operation can be defined as:
\begin{equation}
  m^{'}(A) = \begin{cases}
    \alpha \times m(A) & \forall A \subset \Theta,\\
    1 - \alpha + \alpha \times  m(\Theta) & \text{if}~ A = \Theta.
  \end{cases}
\end{equation}
If $\alpha = 1$, the evidence is completely reliable and the BBA will remain unchanged. On the contrary, if $\alpha=0$, the evidence is completely
unreliable. In this case the so-called vacuous belief function, $m(\Theta)=1$ can be got. It describes our total ignorance.

Before evoking the discounting process, the reliability of each sources should be known.
One possible way to estimate the reliability  is to use  confusion matrices~\cite{martin2005comparative}. 
Generally, the goal of discounting  is to reduce global conflict before combination. One can assume that
the conflict comes from the unreliability of the sources. Therefore, the source reliability estimation  is to some extent
linked to the estimation of conflict between sources.

%

Hence, Martin {\em et al.} proposed to use a conflict measure to evaluate  the relative reliability of experts \cite{martin2008conflict}.
Once the degree of conflict is computed, the relative reliability of the source can be computed accordingly. Suppose there are $S$ sources, $\mathcal{S}=\{s_1,s_2,\cdots,s_S\}$, the  reliability discounting factor $\alpha_j$ of source $s_j$ can be defined as follows:
\begin{equation} \label{reliablitydiscount}
  \alpha_j = f\left(\conf\left(s_j, \mathcal{S}\right)\right),
\end{equation}
where 
$\conf\left(s_j, \mathcal{S}\right)$ quantifies the degree that source $s_j$ conflicts with the other sources in $\mathcal{S}$, and $f$ is a decreasing function. The following function is suggested by the authors:
\begin{equation}
  \alpha_j = \left(1-\conf\left(s_j, \mathcal{S}\right)^\lambda\right)^{\frac{1}{\lambda}},
\end{equation}
where $\lambda > 0$.

\subsection{Simple support functions}
Suppose $m$ is a BBA defined on the frame of discernment $\Theta$. If there exists a subset $A \subseteq \Theta$ such that $m$ could be expressed in the following form:
\begin{equation}
  m(X) = \begin{cases}
    w  & X = \Theta, \\
    1 - w & X = A, \\
    0 & \text{otherwise}.
  \end{cases}
\end{equation}
where $w \in \left[0,1\right]$, then the belief function related to BBA $m$ is called a Simple Support Function (SSF) (also called simple mass function) \cite{ds2} focused on $A$. Such a SSF can be denoted by $A^w(\cdot)$ where the exponent $w$ of the focal element $A$ is the basic belief mass (bbm)  given to the frame of discernment $\Theta$, $m(\Theta)$. The complement of $w$ to 1, {\em i.e.} $1-w$, is the bbm allocated to $A$ \cite{smets1995canonical}. If $w=1$ the mass function represents the total ignorance, if $w=0$ the mass function is a categorical BBA on $A$.

A belief function is separable if it is a SSF or if it is the conjunctive combination of some SSFs \cite{denoeux2008conjunctive}. In the work of \cite{denoeux2008conjunctive}, this kind of separable masses is called u-separable where ``u" stands for ``unnormalized", indicating the conjunctive rule is the unnormalized version of Dempster-Shafer rule.
The set of separable mass functions is not obvious
to obtain.
It is easy to see consonant mass functions (the focal elements are nested) are separable.
Smets \citet{smets1995canonical} defined the Generalized Simple Support Function (GSSF) by relaxing the weight $w$ to $[0,\infty)$.
Those GSSFs with $w\in (1,\infty)$ are called Inverse Simple Support Functions (ISSF). He proved all non-dogmatic mass functions are separable if one uses GSSFs. 

\section{A combination rule for a large number of mass functions}

The conjunctive combination rule tries to reinforce the belief on the focal elements with which most of the sources agree.
However, in this rule, the empty set is an absorbing element. When combing
inconsistent BBAs, the mass assigned to the empty set
tends quickly to 1 with the increasing number of sources~\cite{martin2008conflict}.
Consequently, when using Dempster rule, the gap between $\kappa$ and
1 may rapidly exceed machine precision, even if the combination is valid theoretically.
In that case the fused BBAs by the conjunctive rules  (normalized or not) and the pignistic probability are inefficient due to
the limitation of machine precision. Moreover, the conjunctive combination rule assumes that  all the sources are reliable,
which is difficult to reach or to verify in real applications.

In the theory of belief functions, the idea to the reinforce belief and the alarm role of the empty set in the conjunctive rule are essential.
In order to propose a rule which can be applicable when the number  of mass functions to combine is large and keep
the previous behavior, the following assumptions are made:
\begin{itemize}
\item The majority of sources are reliable;
 \item The larger extent one source is consistent with others, the more reliable the source is;
 \item The sources are cognitively  independent \cite{smets1993belief}.
\end{itemize}

Based on these assumptions, the proposed rule will discount the mass functions according to the
number of sources providing BBAs with the same focal elements. The discounting factor is directly given by the proportion of
mass functions with the same focal elements. As a result, the rule will give more
importance to the groups of mass functions 
that are in a domain, and it is free of auto-conflict~\cite{Martin06a}.
This procedure can be used to  elicit of the majority opinion.

The simple support mass functions are considered here. In this case,  the mass functions can be grouped in the light of their focal
elements (except the frame $\Theta$). To make the rule applicable on separable mass functions, 
the decomposition process should be performed to decompose  each BBA into simple
support mass functions.  In most of applications, the basic belief can be defined using separable mass functions, such as simple support
functions~\cite{denoeux1995k} 
and consonant mass functions~\cite{Dubois90a}.

Hereafter we describe the proposed $\LNS$ rule for simple support functions, and then an approximation calculation method
of $\LNS$ rule is suggested.

\subsection{Combination of many simple support functions}
Suppose that each evidence  is represented by a SSF. Then all the BBAs  can be divided into at most $2^n$ groups (where $n=|\Theta|$). It is easy to
see that there is no conflict at all in each group because of consistency. The focal elements of the SSF are singletons  and
$\Theta$ itself. For the combination of BBAs inside each group, the conjunctive rule can be employed directly. Then the fused BBAs are  discounted
according to the number of mass functions in each group. Finally, the global combination of the BBAs of different groups is preformed also using the
conjunctive rule. Suppose  that all BBAs are defined on the frame of
discernment $$\Theta=\{\theta_1,\theta_2,\cdots,\theta_n\},$$ and denoted by  $$m_j = (A_i)^{w_j}, j=1,\cdots, S, i=1,2,\cdots,c,$$ where $c\leq 2^n$. 
The detailed process of the combination is listed as follows. Our proposed rule called $\LNS$ for Large Number of Sources rule is composed of the four
following steps:
\begin{enumerate}
  \item Cluster the simple BBAs into $c$ groups based on their focal element $A_i$. For the convenience, each class is labeled by
  its corresponding focal element.
  \item Combine the BBAs in the same group. Denote the combined BBA in group $A_k$ by SSF \linebreak $\hat{m}_k= (A_k)^{\hat{w}_k}$, $k=1,2,\cdots,c$. For the conjunctive combination rule we have:
  \begin{equation}
   \hat{m}_k=\ocap_{j=1, \cdots, s_k}m_j=(A_k)^{\displaystyle \prod_{j=1}^{s _k} w_j}
  \end{equation}
where the number of BBAs in group $A_k$ is $s_k$. In order to consider the total ignorance as a neutral element of the rule, if $A_k=\Theta$ we allow
$s_k=0$. 
  \item Reliability-based discounting. Suppose the fused BBA of all the mass functions
  in $A_k$ be $\hat{m}_k$. At this time, each group can be regarded as a source, and there are $c$ sources in total. The reliability of one source can be estimated by comparing to the group of sources. In our opinion, the reliability of source $A_k$ is related to the proportion
  of BBAs in  this group. The larger the number of BBAs in group $A_k$ is, the more reliable $A_k$ is.  Then the reliability discounting
  factor of $\hat{m}_k$, denoted by $\alpha_k$, can be defined as:
      \begin{equation}
      \label{discountfactorSimple}
      \alpha_k = \frac{s_k}{\displaystyle  \sum_{i=1}^{c} s_i}.
      \end{equation}
      Another version of the  discounting factor can be determined  by a factor taking into account the precision of the group:
      \begin{equation}
      \label{discountfactor}
      \alpha_k = \frac{\beta_k^\eta s_k}{\displaystyle \sum_{i=1}^c \beta_i^\eta s_i},
      \end{equation}
      where
      \begin{equation}
        \beta_k = \frac{|\Theta|}{|A_k|}.
      \end{equation}

      Parameter $\eta$ can be used to adjust the precision of the combination results.  The larger the value of $\eta$ is, the
      less imprecise the resulting BBA is.
      The discounted BBA of $\hat{m}_k$ can be denoted by SSF \linebreak $\hat{m}_k^{'}=(A_k)^{\hat{w}_k^{'}}$ with $\hat{w}_k^{'} = 1 - \alpha_k + \alpha_k \hat{w}_k$. As we can see, when the number of BBAs in one group is larger, $\alpha$ is closer to 1. That is to say, the fused mass in this group is more reliable.

  \item Global combine the fused BBAs in  different groups using the conjunctive rule:
  \begin{equation}
  \label{lastcombination}
   m_\LNS=\ocap_{k=1, \cdots, c} \hat{m}_k'=\ocap_{k=1, \cdots, c} (A_k)^{\hat{w}_k^{'}}
  \end{equation}

\end{enumerate}
The previous mentioned methods in Section~\ref{reliability} to estimate reliability are much more complex than the proposed method here. Indeed, usually the distance between BBAs should be calculated or a special learning process is required. In $\LNS$ rule, to evaluate the reliability discounting factor, we only need to count the number of SSFs in each group. But other reliability estimation methods can also be used.

In the last step of combination, as the number of mass functions that takes part in the global combination is small (at most $2^n$), other combination rules such as $\DP$ rule \cite{dubois1988representation} and PCR  rules \citet{Martin06a} are also possible in practice instead of Eq.~\eqref{lastcombination}.

\subsection{LNS properties}
\label{secproperties}
The proposed rule is commutative, but not associative. The rule is not idempotent, but there is no absorbing element. 
The vacuous mass function is a neutral element of the $\LNS$ rule. 

There are four steps when applying $\LNS$ rule\footnote{The source code for $\LNS$ rule could be found in R package \textit{ibelief} \cite{ibelief}.}:
decomposition (not necessary for simple support mass functions), inner-group combination, discounting and global combination. The $\LNS$ rule has the
same memory complexity as some other rules such as conjunctive, $\DS$ and cautious rules if all the rules are combined globally
using FMT method. Only $\DP$ and $\PCR$ rules have higher memory complexity because of the partial conflict to manage. Suppose the number of
mass functions to combine is $S$, and the number of elements in the frame of discernment  is $n$. The complexity for decomposing\footnote{In the
decomposing process, the Fast M{\"o}buis Transform method is used.
 }
mass functions to SSFs is $O(Sn2^n)$.
For combining the mass functions in each group, due to the structure of the simple support mass functions, we only need to
calculate the product of the masses on only one
focal element $\Theta$. Thus the complexity is $O(S)$. The complexity of the discounting is $O(2^n)$. In the
process of global combination, the BBAs are all SSFs. If we use  the Fast M{\"o}buis Transform method
, the complexity is $O(n2^n)$. Moreover there are at most $2^n$ mass functions participating  the following discounting and global
conjunctive combination processes. Since in most application cases with a large number of mass functions, we have $2^n \ll S$, the last  two steps are
not very time-consuming. The total complexity of $\LNS$ is $O(Sn2^n+S+2^n+n2^n)$ and so is equivalent to $O(Sn2^n)$.

We remark here that one of the assumptions of $\LNS$ rule is that the majority of sources are reliable.  However, this condition  is
not always satisfied in every applicative context. Consider here an example with two sensor
technologies: TA and TB. The system has two TA-sensors ($S_1$ and $S_2$),  and one TB-sensor $S_3$. Suppose also a
parasite signal causes TA sensors to malfunction. In this situation, the majority of sensors are unreliable, and we could not
get a good result if the $\LNS$ rule is used directly as $\LNS(S_1,S_2,S_3)$ at this time.
Actually there is an underlying hierarchy in the sources of information,  $\LNS$
rule could be evoked according to  the hierarchy, such as $\LNS(\LNS(S_1,S_2),S_3)$. We will study that more in the future work.

\vspace{-0.6em}
\section{Experiments}
\label{Experiments}
\vspace{-0.8em}
To illustrate the behavior of the proposed combination rule $\LNS$ and to compare with other classical rules, several experiments will be conducted here. 
Some different types of randomly generated mass functions  will be used. The function \textit{RandomMass} in R package \textit{ibelief}
\cite{ibelief} is adopted to generate random mass functions.

\Exp.  In the crowdsourcing applications, all the
users can provide some imprecise and uncertain answers. But only a few are trusty.
The elicitation of the majority opinion is very important to identify the experts.
Assume that the answers by different uses are in the form of the mass
functions over the same discernment frame $\Theta=\{\theta_1,\theta_2,\theta_3\},$ denoted by $m_j, j = 1, 2, \cdots, 6$. The assignments
of all the mass functions are as follows:
\begin{align*}
& m_1: m_1(\{\theta_2\}) = 0.9, ~m_1(\Theta)=0.1,\nonumber \\
& m_2: m_2(\{\theta_1\}) = 0.1, ~m_2(\Theta)=0.9,\nonumber \\
& m_3: m_3(\{\theta_1\}) = 0.2, ~m_3(\Theta)=0.8,\nonumber \\
& m_4: m_4(\{\theta_1\}) = 0.3, ~m_4(\Theta)=0.7,\nonumber\\
& m_5: m_5(\{\theta_1\}) = 0.1, ~m_5(\Theta)=0.9,\nonumber\\
& m_6: m_6(\{\theta_1\}) = 0.2, ~m_6(\Theta)=0.8. \nonumber\\
\end{align*}
    As can be seen from the above equations, there are five out of six mass functions ($m_2,m_3, m_4, m_5, m_6$) assigning a large value
on $\theta_1$, while $m_1$ delivers a function strongly committed to another solution.
It indicates that the first source is obviously different from the other five sources.

The combination results by conjunctive rule, $\DS$ rule, disjunctive rule, $\DP$ rule, $\PCR$ rule, cautious rule, average rule
and the proposed $\LNS$ rule\footnote{As the focal elements are singletons except $\Theta$, parameter $\eta$ has no effects on
the final results by the proposed combination rule.} are depicted in Table
\ref{6massEx1}. As can be observed, the conjunctive rule assigns most of the belief to
the empty set, regarding the sources as highly conflictual.
$\DS$ rule, $\DP$ rule, $\PCR$ rule and average rule redistribute
all the global conflict to other focal elements. Disjunctive rule gives the total ignorance mass functions.
Cautious rule and the proposed  $\LNS$ rule keep some of the conflict and redistribute the remaining. From the original six BBAs,
we can see that there are five mass functions supporting $\{\theta_1\}$, while only one supporting $\{\theta_2\}$. The six mass functions
are not conflicting, because the majority of evidence shows the preference of $\{\theta_1\}$. We consider
here source 1 is not reliable since it contradicts with all the other sources. But the belief
given to $\{\theta_2\}$ is more than that to $\{\theta_1\}$
when using  $\DS$, $\DP$, $\PCR$, and the cautious rule, which indicates that these rules are not robust to the unreliable evidence.
The result by the average rule gives equal evidence to $\{\theta_1\}$ and $\{\theta_2\}$.
The obtained fused BBA by the proposed rule
assigns  the largest  mass to focal element $\{\theta_1\}$, which is
consistent with the intuition. It keeps a certain level of global conflict, and  at
the same time reflects the superiority of $\{\theta_1\}$ compared with $\{\theta_2\}$. From the results we can see that only
the $\LNS$ rule can correctly elicit the major opinion.

\begin{table*}[ht]
\centering \caption{The combination of six masses. For the names of columns, $\theta_{ij}$ is used to denote $\{\theta_i, \theta_j\}$.}
\resizebox{\textwidth}{!}{
\begin{tabular}{rrrrrrrrr}
  \hline
 & Conjunctive & DS & Disjunctive & DP & PCR6 & Cautious & Average & $\LNS$ \\
  \hline
$\emptyset$ & 0.57341 & 0.00000 & 0.00000 & 0.00000 & 0.00000 & 0.27000 & 0.00000 & 0.07964 \\
  $\theta_1$ & 0.06371 & 0.14935 & 0.00000 & 0.06371 & 0.10644 & 0.03000 & 0.15000 & 0.45129 \\
  $\theta_2$ & 0.32659 & 0.76558 & 0.00000 & 0.32659 & 0.45139 & 0.63000 & 0.15000 & 0.07036 \\
  $\theta_{12}$ & 0.00000 & 0.00000 & 0.00011 & 0.08165 & 0.00000 & 0.00000 & 0.00000 & 0.00000 \\
  $\theta_3$ & 0.00000 & 0.00000 & 0.00000 & 0.00000 & 0.00000 & 0.00000 & 0.00000 & 0.00000 \\
  $\theta_{13}$ & 0.00000 & 0.00000 & 0.00000 & 0.00000 & 0.00000 & 0.00000 & 0.00000 & 0.00000 \\
  $\theta_{23}$ & 0.00000 & 0.00000 & 0.00000 & 0.00000 & 0.00000 & 0.00000 & 0.00000 & 0.00000 \\
  $\Theta$ & 0.03629 & 0.08506 & 0.99989 & 0.03629 & 0.44217 & 0.07000 & 0.70000 & 0.39871 \\
   \hline
\end{tabular}}
\label{6massEx1}
\end{table*}

\begin{center} \begin{figure}[!thbt] \centering
		\includegraphics[width=0.9\linewidth]{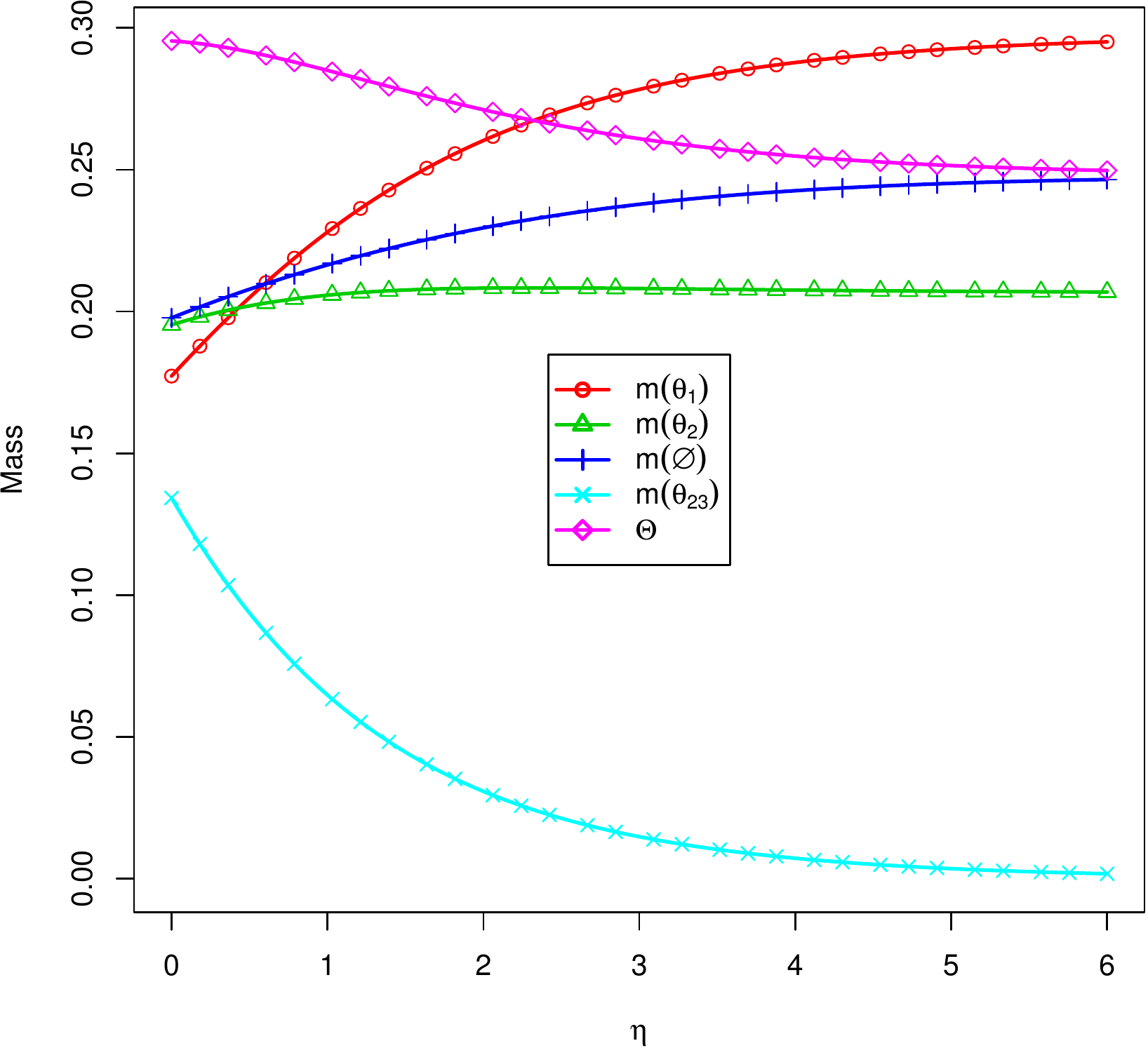}
\caption{The BBA after the combination  for three types of SSFs using $\LNS$ rule.  The mass functions
are generated randomly, and $\LNS$ rule is evoked with different values of $\eta$ ranging from 0 to 6.} \label{Exp5witheta1} \end{figure} \end{center}

\begin{center} \begin{figure}[!thbt] \centering
       \includegraphics[width=0.9\linewidth]{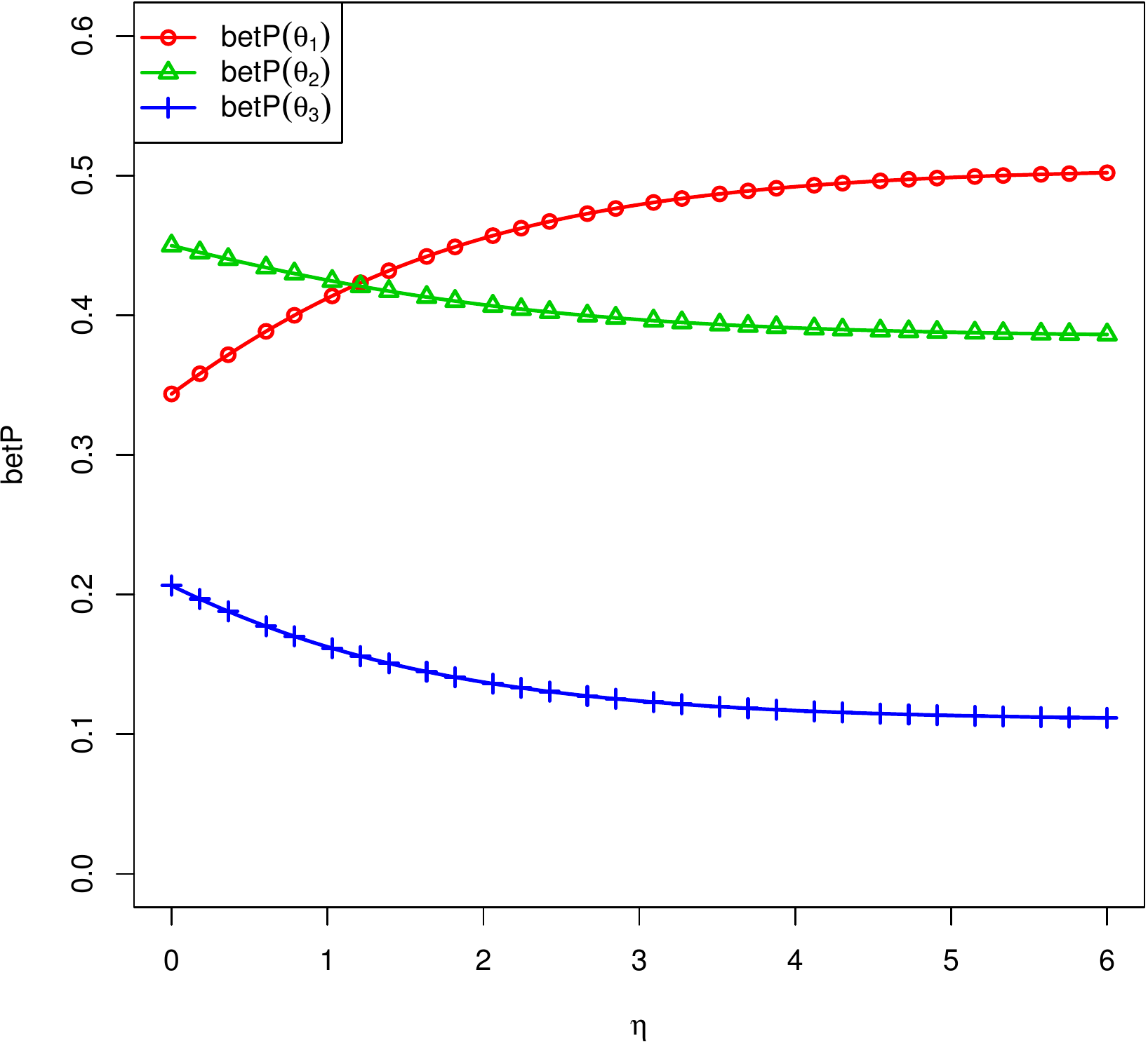} \hfill
	\caption{The Pignistic  probability after the Combination  for three types of SSFs using $\LNS$ rule.  The mass functions
are generated randomly, and $\LNS$ rule is evoked with different values of $\eta$ ranging from 0 to 6.} \label{Exp5witheta2} \end{figure} \end{center}

\vspace{1em}
\Exp. We test here the influence of parameters $\eta$ and $\beta$ in $\LNS$ rule. Simple support mass functions are
utilized in this experiment. Suppose that the discernment frame under consideration is $\Theta=\{\theta_1,\theta_2,\theta_3\}$. Three types
of SSFs are adopted. First $s_1 = 60$ and $s_2 = 50$ SSFs with focal element $\{\theta_1\}$ and $\{\theta_2\}$ respectively (the other focal element is
$\Theta$) are uniformly generated, and then
$s_3 = 50$ SSFs with focal element $\theta_{23}\triangleq  \{\theta_2, \theta_3\}$ are generated. The value of masses are randomly generated. Different values of $\eta$
ranging from 0 to 6 are used to test. The
mass values in the fused BBA by $\LNS$ varying with $\eta$ are displayed in Figure \ref{Exp5witheta1}, and
the corresponding pignistic probabilities
are shown in Figure \ref{Exp5witheta2}.

From these figures, we can see that $\eta$ can have some effects on the final decision.  Figure \ref{Exp5witheta1} shows that
with the increasing of $\beta$, the mass function assigned to the singleton focal elements increases. On the contrary, the mass given to the focal
element whose cardinality is bigger than one decreases.
In fact parameter $\beta$ in $\LNS$ aims at  weakening the imprecise
evidence which gives only positive mass to focal elements with high cardinality, and
the exponent $\eta$ allows  to control the degree of discounting. If $\eta$ is larger, we give more
weights to the  sources of evidence whose focal elements are more specific, and more discount will be committed to the imprecise
evidence. As a result, in the experiment when $\eta$ is larger
than 1.2, $\BetP(\theta_1) > \BetP(\theta_2)$ (Figure \ref{Exp5witheta2}). At this time
the mass functions with focal element $\{\theta_2, \theta_3\}$ make little contribution
to the fusion process, while the final decision mainly depends on
the other two types of simple support mass functions with singletons as focal elements.

In real applications, $\eta$ could be determined based on  specific requirement. This work is not specially focusing on
how to determine $\eta$, thus in the following experiment we will set $\eta=1$ as default.

\vspace{0.2em}
\Exp. 
The goal of this experiment is to show how Dempster's degree of conflict is dealt with by  most of rules when combining a large number of conflicting sources.

In this experiment, the frame  of discernment is set to $\Theta = \{ \theta_1, \theta_2\}$. Assume that there are only 2 focal elements on each BBA.
One is the whole frame $\Theta$, and the other is any of  the singletons ($\{\theta_1\}$ or $\{\theta_2\}$). The number of BBAs which have the focal
element $\{\theta_1\}$ is set to $s_1$, while that with $\{\theta_2\}$ is $s_2$. We fix
the value of $s_2$, and let $s_1 = t * n_2$, with $t$ a positive integer. We generate $S = s_1 + s_2$ such kind of BBAs randomly, but only withholding
the BBAs for which the mass value assigned to $\{\theta_1\}$ or $\{\theta_2\}$ is greater than 0.5.

\begin{center} \begin{figure}[!thbt] \centering
		\includegraphics[width=0.9\linewidth]{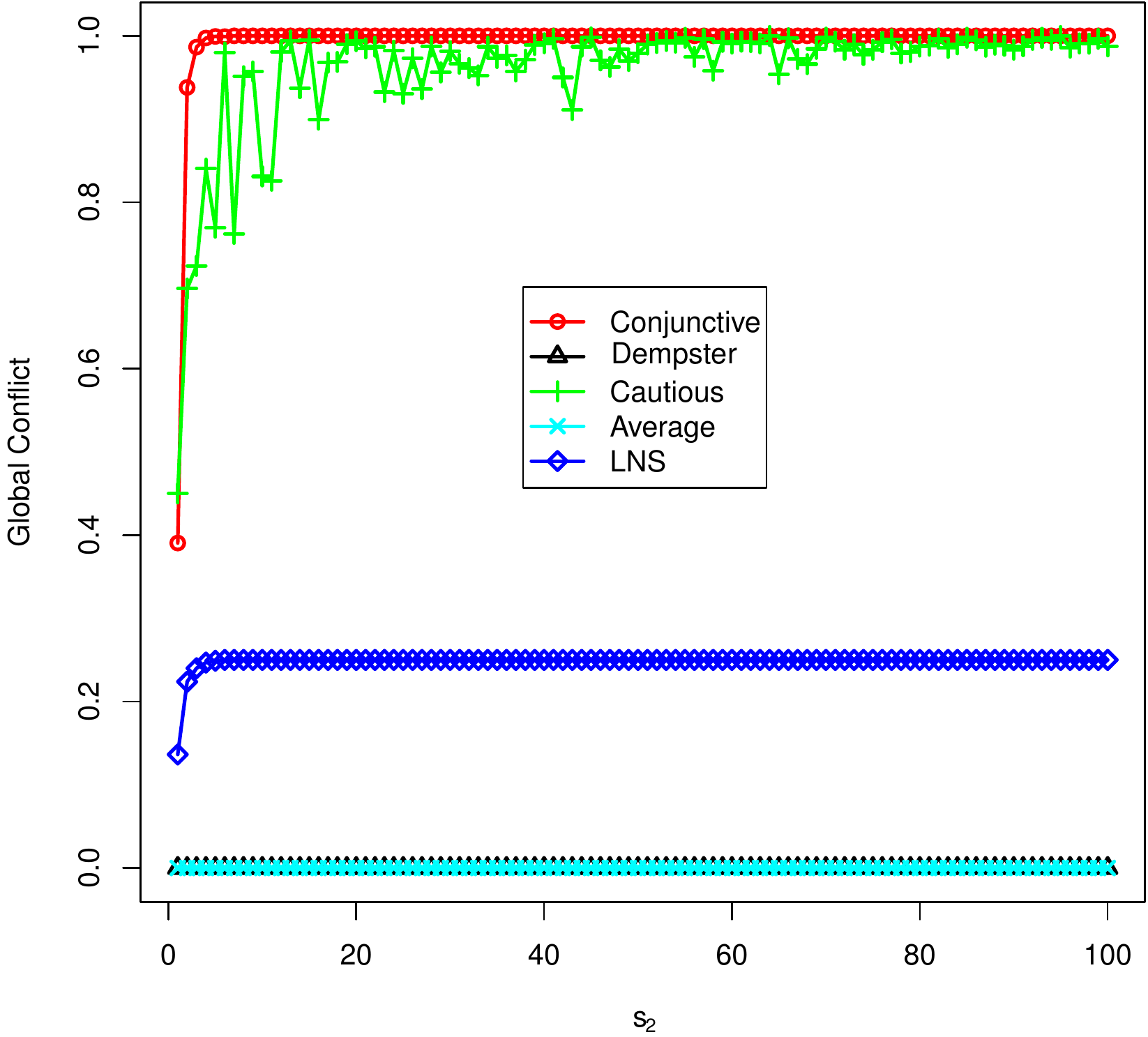}\hfill
\caption{The global conflict after the combination with $s_1 = s_2$.} \label{confwithn1} \end{figure} \end{center}

\begin{center} \begin{figure}[!thbt] \centering
        \includegraphics[width=0.9\linewidth]{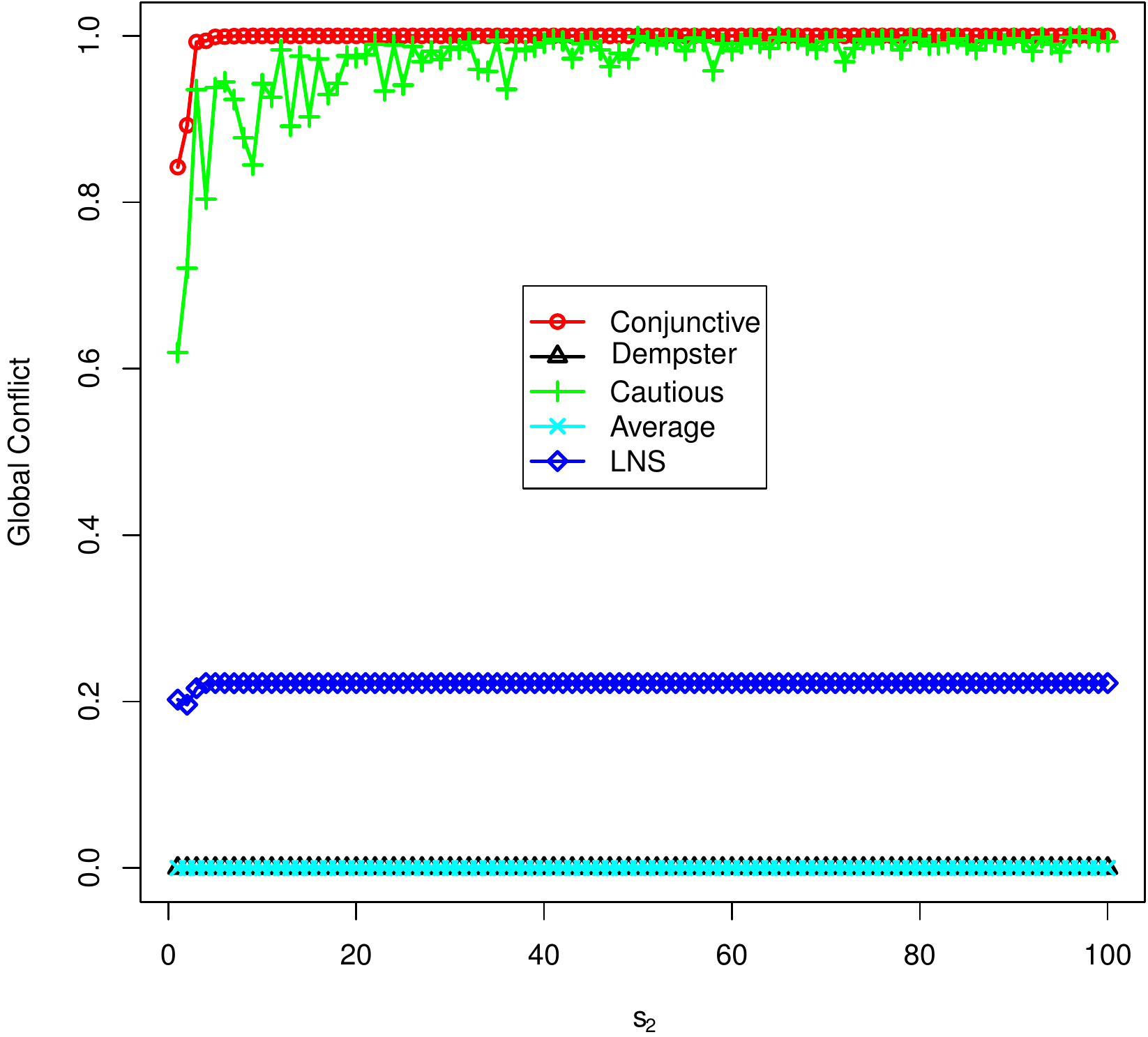} \hfill
\caption{The global conflict after the combination with $s_1 = 2*s_2$.} \label{confwithn2} \end{figure} \end{center}

\begin{center} \begin{figure}[!thbt] \centering
	\includegraphics[width=0.9\linewidth]{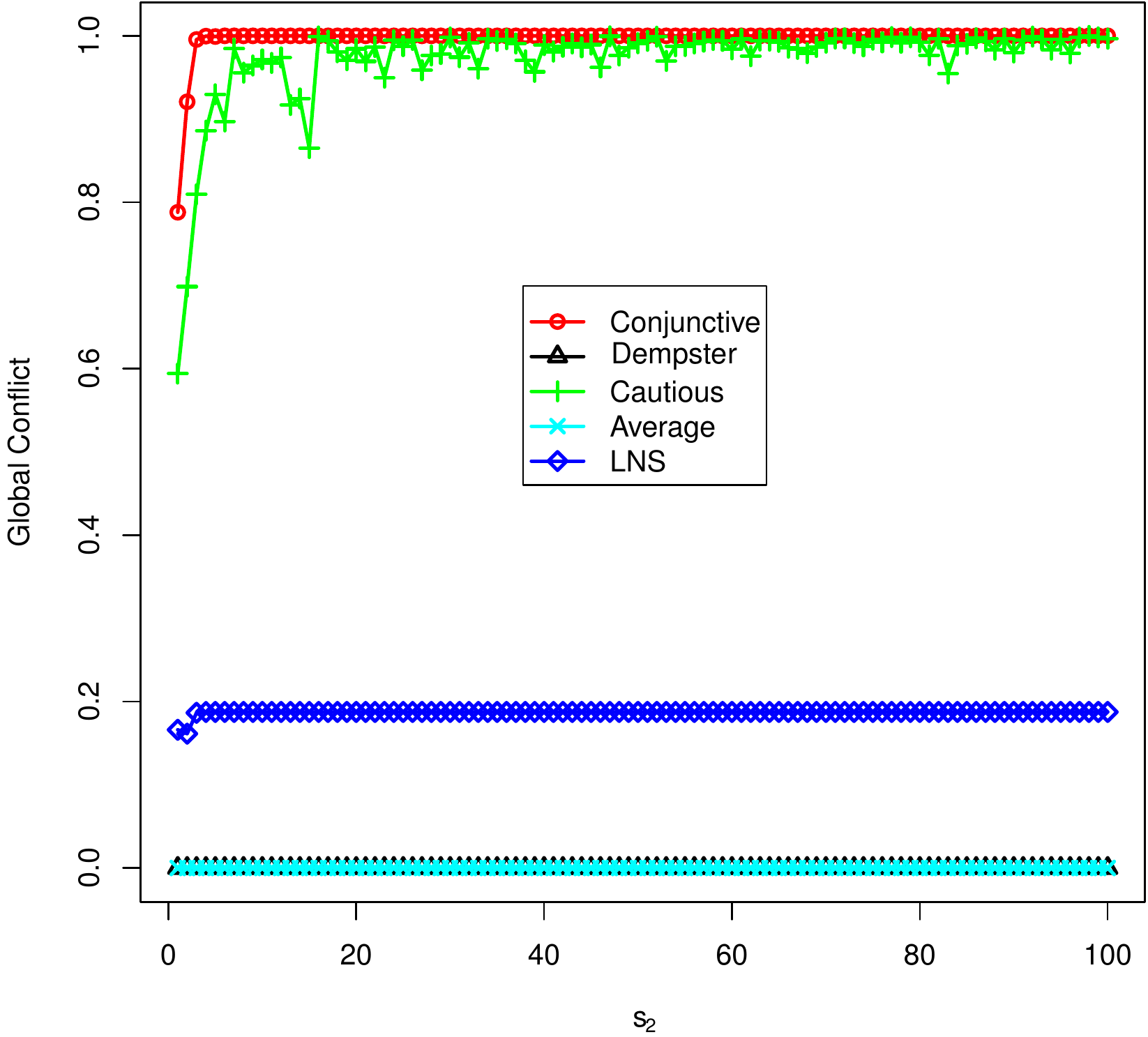}\hfill
\caption{The global conflict after the combination with $s_1 = 3 * s_2$.} \label{confwithn3} \end{figure} \end{center}

\begin{center} \begin{figure}[!thbt] \centering
        \includegraphics[width=0.9\linewidth]{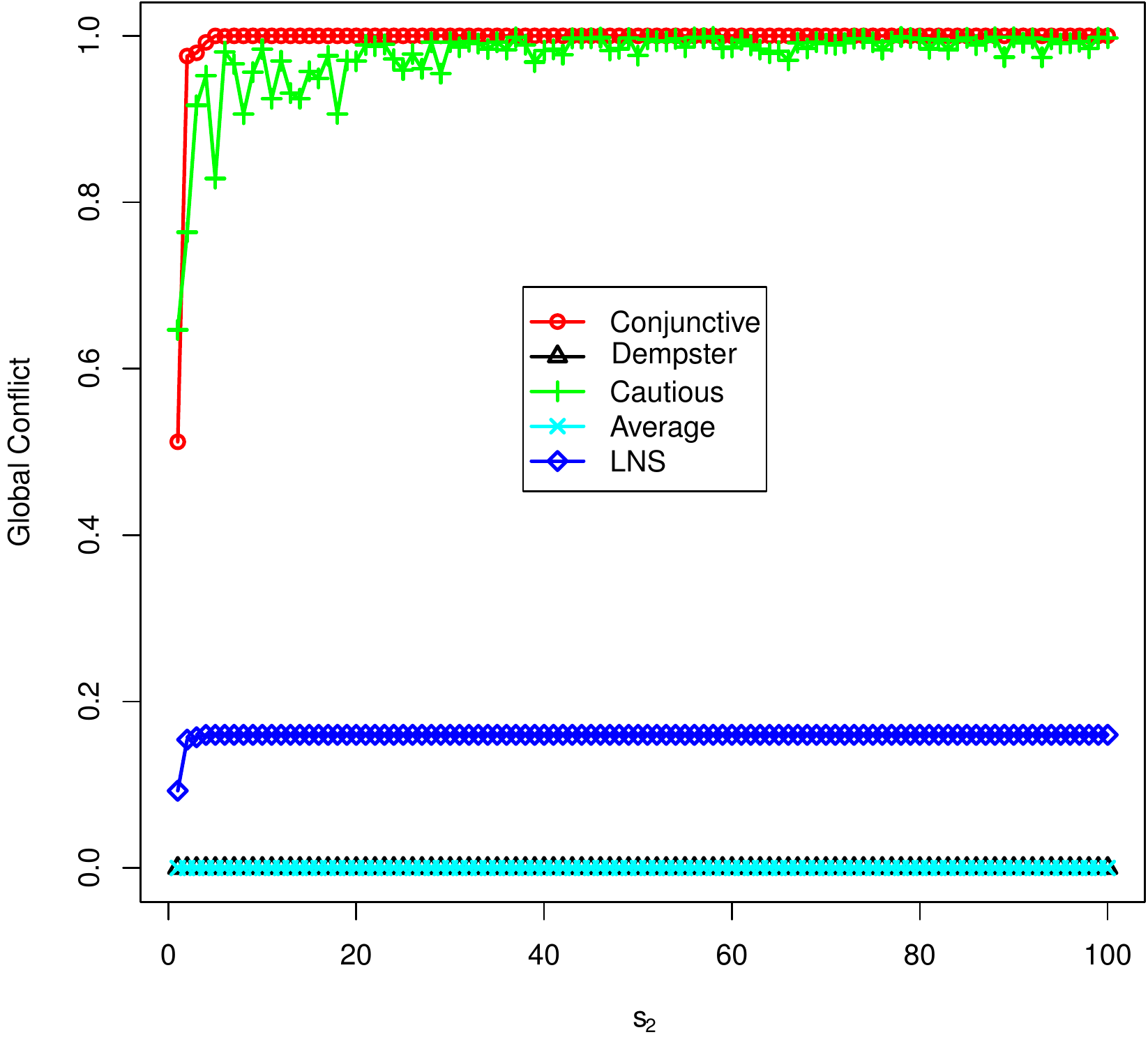} \hfill
\caption{The global conflict after the combination with  $s_1 = 4 *s_2$.} \label{confwithn4} \end{figure} \end{center}

\begin{center} \begin{figure}[!thbt] \centering
		\includegraphics[width=0.9\linewidth]{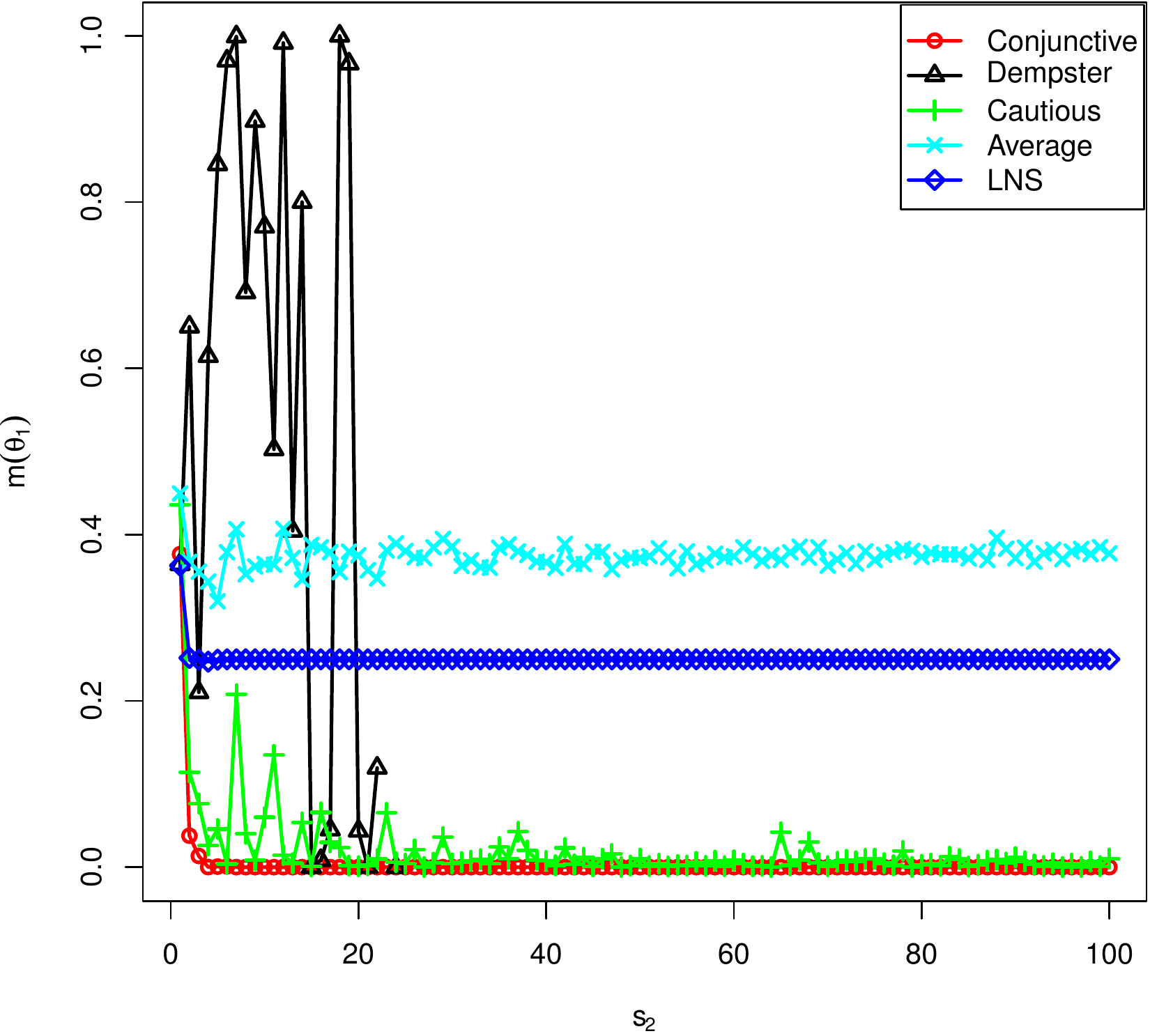}\hfill
\caption{The mass on  $\{\theta_1\}$ after the combination with  $s_1 = s_2$.} \label{theta1withn1} \end{figure} \end{center}

\begin{center} \begin{figure}[!thbt] \centering
        \includegraphics[width=0.9\linewidth]{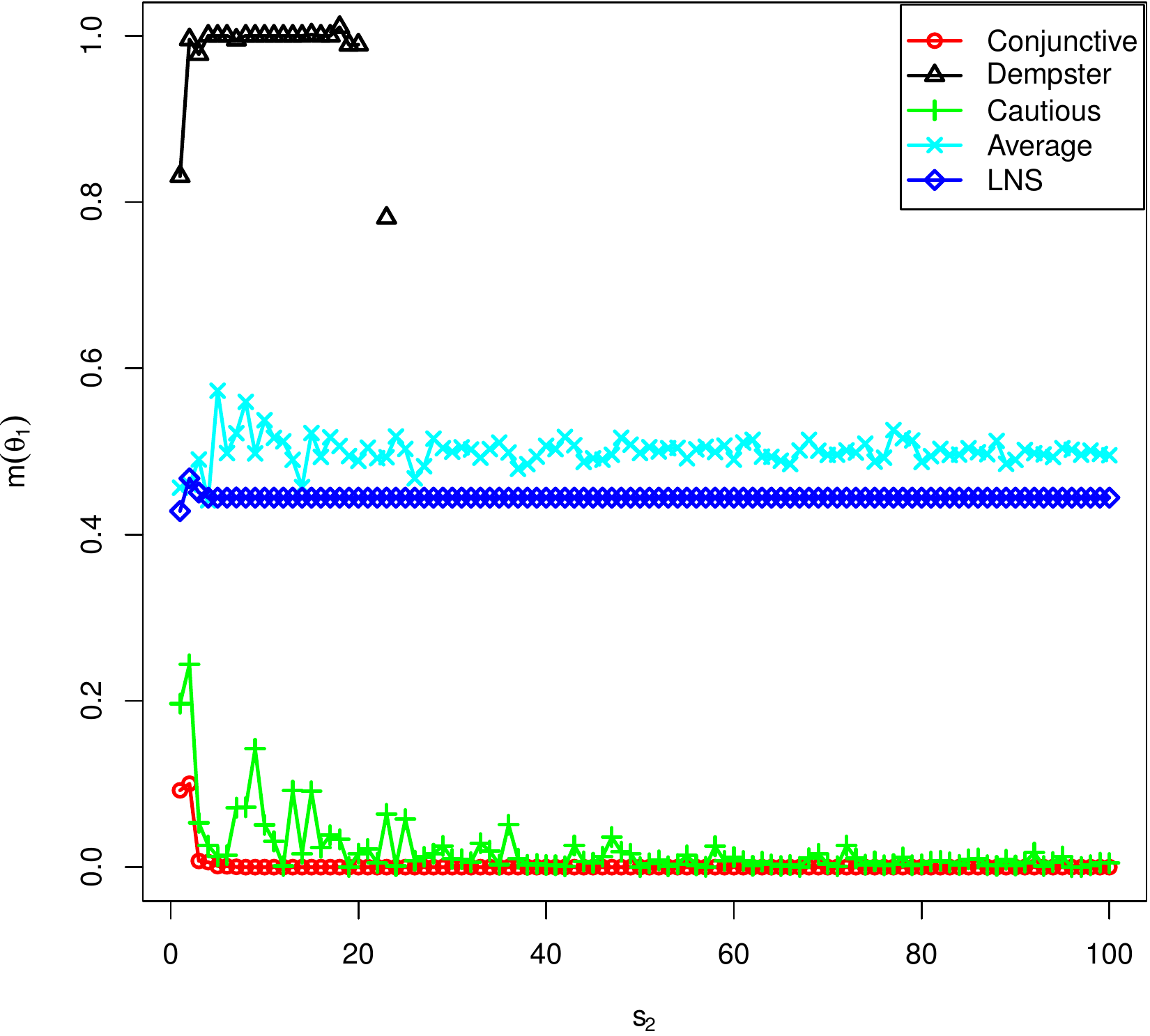} \hfill
\caption{The mass on  $\{\theta_1\}$ after the combination with $s_1 = 2*s_2$.} \label{theta1withn2} \end{figure} \end{center}

\begin{center} \begin{figure}[!thbt] \centering
	\includegraphics[width=0.9\linewidth]{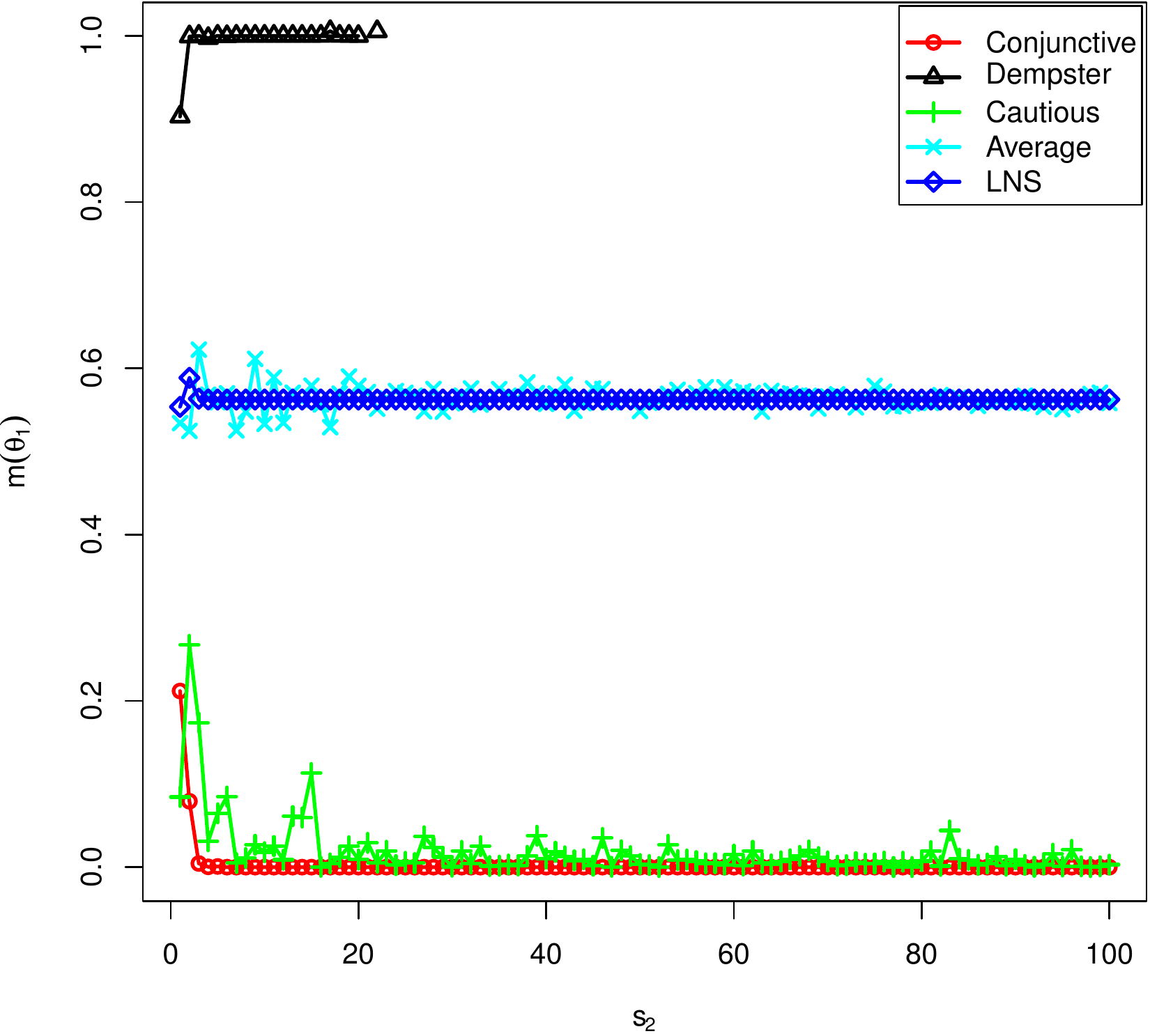}\hfill
\caption{The mass on  $\{\theta_1\}$ after the combination with $s_1 = 3 * s_2$.} \label{theta1withn3} \end{figure} \end{center}

\begin{center} \begin{figure}[!thbt] \centering
        \includegraphics[width=0.9\linewidth]{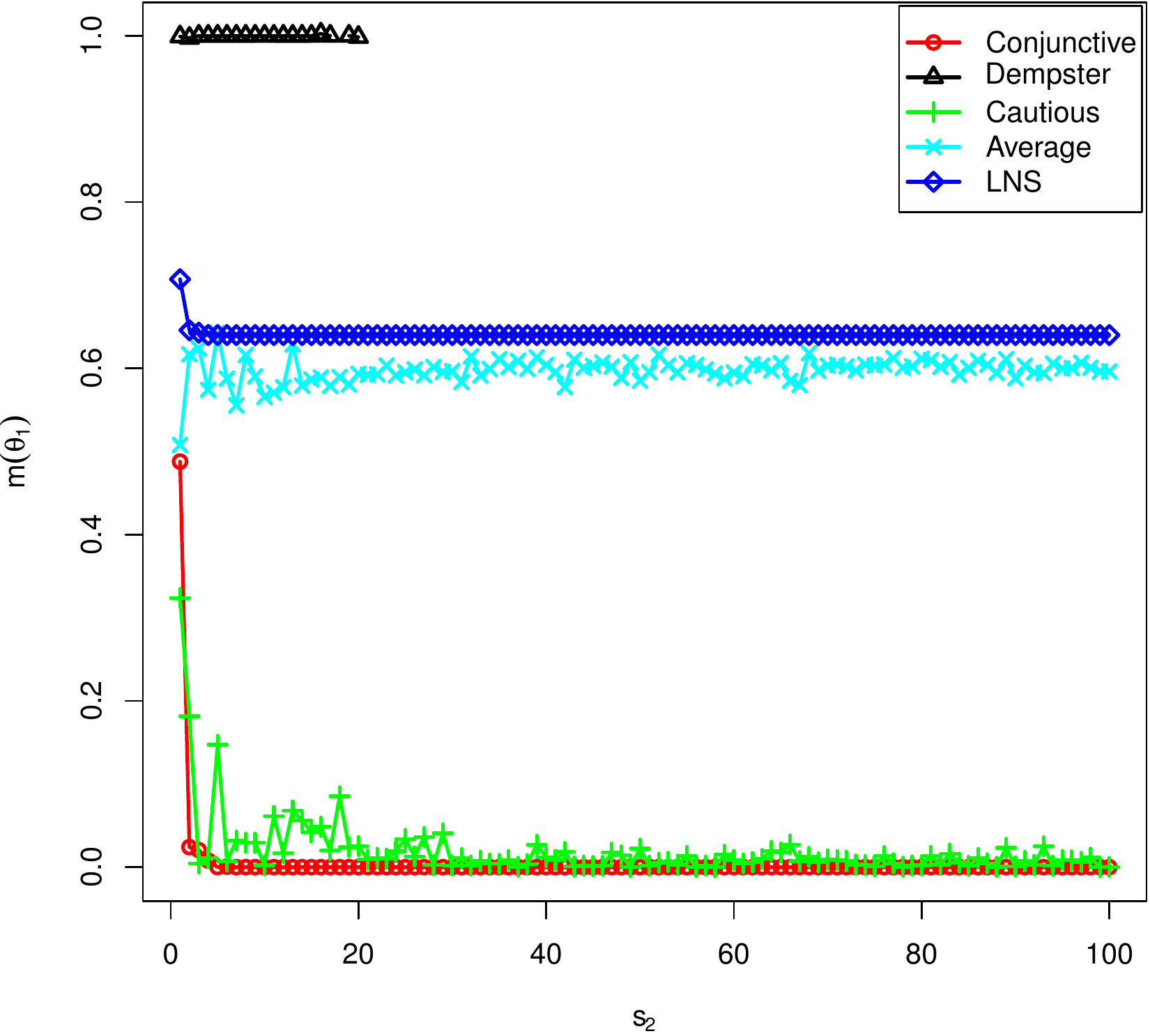} \hfill
\caption{The mass on  $\{\theta_1\}$ after the combination with $s_1 = 4 *s_2$.} \label{theta1withn4} \end{figure} \end{center}

Four values of $t$ are considered here: $t=1, 2, 3, 4$. If $t=1$, $s_1 = s _2 = S/2$.  If $t=2$, the number of mass functions
supporting $\{\theta_1\}$ is two
times of that supporting $\{\theta_2\}$, and so on.   The global conflict (mass given to the empty set) after the combination with different values of $s_2$ for the four cases is displayed in Figures \ref{confwithn1}--\ref{confwithn4} respectively. The mass assigned to the focal element $\{\theta_1\}$ with different combination approaches is shown in Figures \ref{theta1withn1}--\ref{theta1withn4}.

\vspace{-6em}
It is intuitive that when $t$ becomes larger, the global conflict should be smaller and we should give more belief to the focal element $\{\theta_1\}$. From  Figures~\ref{confwithn1}--\ref{theta1withn4} we can see that only the results by $\LNS$ rule are in accordance with this common sense.  The simple average rule assigns larger BBA to $\{\theta_1\}$, but it does not keep any conflict. $\DS$ rule could not work at all when $s_2$ is larger than 20\footnote{In Figures \ref{theta1withn1}--\ref{theta1withn4}, the mass given to $\{\theta_1\}$ by $\DS$ rule is not displayed when $S$ is large (and also for some small $S$), because in these cases the global conflict is 1 and the normalization could not be processed.}, as it regards these BBAs as highly conflict. Although the conjunctive rule and cautious rule could work when combining a larger number of mass functions,  the
obtained fused mass function is $m(\emptyset)\approx 1$, which is useless for decision in practical.

From Figures \ref{confwithn1}--\ref{confwithn4}, we can see a kind of limit of the global conflict for the $\LNS$ rule. In fact, the mass on the empty set for this rule is
also depending on the size of the frame of discernment and more directly on the number of groups created in the first step of the rule. The limit
value of the global conflict will tend to 1 with the increase of the size of discernment when considering only categorical BBAs on different
singletons.

\section{Conclusion}
There is usually a lot of uncertain information in big data applications. The theory of belief functions is
a flexile framework to deal with imprecise and uncertain information, especially it
provides many ways for the task of information fusion. However, although lots of
combination rules have been designed in  recent years in this framework, most of them are not applicable when the
number of source to combine is quite large due to the complexity or the existing absorbing element.

We propose here a new combination rule, named $\LNS$ rule,  preserving the principle of the conjunctive rule.
This rule first groups the mass functions according to their set of focal elements (without auto-conflict). After the
inner combination, the mass functions in each
group can be summarized by one mass function. The reliability of the source is estimated by the proportion of BBAs
in one group. Therefore, after discounting the mass function of each group by the reliability factor, the final combination can be proceeded  by the conjunctive rule (or another rule according to the application).

The $\LNS$ rule  is able to combine a large number of mass functions. The only existing method applicable for a large number of
sources is the average rule. However, that rule may give more importance to few sources
with a high belief (even if the source is not reliable). It cannot capture the conflict
between the sources. The proposed rule has a reasonable complexity (lower than the $\DP$ and $\PCR$ rules). Moreover,
it can provide reasonable combination results and can be used to elicit the major opinion. This is of practical value in
the crowdsourcing system.

Overall, this work provides a perspective for the application of belief functions on big data. We will study how to apply  $\LNS$ rule
on the problems of social network and crowdsourcing in the future research work.

\section*{Acknowledgements}
This work was supported by the National
Natural Science Foundation of China (Nos.61135001, 61403310, 61672431) and  the
Fundamental Research Funds for the Central Universities of China (No.3102016QD088).

\bibliographystyle{IEEEtran}
\bibliography{paperlist}
\end{document}